\documentclass{article} 
\usepackage{iclr2016_conference,times}
\usepackage{hyperref}
\pdfoutput=1
\usepackage{times}
\usepackage{helvet}
\usepackage{courier}
\usepackage{mathrsfs}
\usepackage{multirow}
\usepackage{amssymb}
\usepackage[cmex10]{amsmath}
\usepackage{paralist}
\usepackage{graphicx}
\usepackage[tight,footnotesize]{subfigure}
\usepackage{threeparttable}
\usepackage{booktabs}
\usepackage{url}
\usepackage{algorithm}

\usepackage{algorithmic}

\graphicspath{{figs/}}

\usepackage{mathptmx}      

\def \g  {\mathbf{g}}

\def \v {\mathbf{v}}
\def \w  {\mathbf{w}}
\def \x {\mathbf{x}}
\def \X {\mathbf{X}}

\DeclareMathOperator*{\argmin}{arg\,min}

\newtheorem{prop}{Proposition}

\def \Mu {\pmb{\mu}}
\def \Nn {\mathcal{N}}
\def \U {\mathcal{U}}
\def \R {\mathbb{R}}

\def \diag{\mathrm{diag}}

\newcommand{\includeMyGraphicX}[1]{\includegraphics[clip=true, height=0.35\textwidth, width=0.45\textwidth]{#1}}

\title{Large-scale Online Feature Selection for Ultra-high Dimensional Sparse Data}

\author{Yue~Wu\\
University of Science and \\
Technology of China\\
\texttt{wye@mail.ustc.edu.cn} \\
\And
Steven C.H. Hoi\thanks{Corresponding author: http://stevenhoi.org/, SIS, SMU, 80 Stamford Road, Singapore 178902}\\
School of Information Systems\\
Singapore Management University, Singapore\\
\texttt{chhoi@smu.edu.sg}
\And
Tao~Mei\\
Microsoft Research, Beijing \\
\texttt{tmei@microsoft.com} \\
\And
Nenghai~Yu\\
University of Science and \\
Technology of China\\
\texttt{ynh@ustc.edu.cn}
}

\begin{document}

\maketitle

\begin{abstract}


Feature selection with large-scale high-dimensional data is important yet very challenging in machine learning and data mining. Online feature selection is a promising new paradigm that is more efficient and scalable than batch feature section methods, but the existing online approaches usually fall short in their inferior efficacy as compared with batch approaches. In this paper, we present a novel second-order online feature selection scheme that is simple yet effective, very fast and extremely scalable to deal with large-scale ultra-high dimensional sparse data streams. The basic idea is to improve the existing first-order online feature selection methods by exploiting second-order information for choosing the subset of important features with high confidence weights. However, unlike many second-order learning methods that often suffer from extra high computational cost, we devise a novel smart algorithm for second-order online feature selection using a MaxHeap-based approach, which is not only more effective than the existing first-order approaches, but also significantly more efficient and scalable for large-scale feature selection with ultra-high dimensional sparse data, as validated from our extensive experiments. Impressively, on a billion-scale synthetic dataset (1-billion dimensions, 1-billion nonzero features, and 1-million samples), our new algorithm took only 8 minutes on a single PC, which is orders of magnitudes faster than traditional batch approaches. \url{http://arxiv.org/abs/1409.7794}

\end{abstract}

\section{Introduction}
In machine learning and data mining, feature selection (FS) is the process of selecting a subset of relevant features and removing irrelevant and redundant features from data towards model construction. It is a very important technique in the era of big data today, and has found applications in a wide range of domains, particularly for scenarios with high-dimensional data. Feature selection has been extensively studied in which various algorithms have been proposed~\citep{liu2005toward}.

Despite the extensive research efforts in literature, most existing feature selection methods are restricted to batch learning settings~\citep{saeys2007review}, which have many critical drawbacks for big data applications. One drawback with batch learning is that they often require the entire training data set to be loaded in memory. This is obviously non-scalable when solving real-world applications with large-scale datasets that exceed memory capacity. Another drawback is that batch learning methods usually assume all training data and their full set of features must be made available prior to the learning task. This assumption does not always hold in many real-world applications where data arrives sequentially (e.g., internet data) and novel features may appear incrementally (e.g., spam email filtering). These drawbacks make traditional batch feature selection techniques non-practical for emerging big data applications.

To overcome the drawbacks of batch feature selection, online feature selection has been explored recently~\citep{perkins2003online,wang2014online,wu2010online}. One state-of-the-art scheme in \citep{wang2014online} attempts to resolve feature selection by exploring online learning techniques. Although it is far more efficient and scalable than batch feature selection techniques, it still falls short in requiring linear time complexity with respect to feature dimensionality and sometimes failing to achieve satisfying learning accuracy when solving difficult tasks.

In this paper, we argue that existing solutions are still not feasible due to high time and memory cost in real world applications with large-scale and ultra-high dimensional data. We propose a simple but smart second order online feature selection algorithm that is extremely efficient, scalable to large scale and ultra-high dimensionality, and effective to address this open challenge. Compared to existing FS methods, the complexity is significantly reduced to be linear to the average number of nonzero features per instance, rather than the full feature dimensionality. In particular, unlike the existing first-order online FS approaches, the proposed algorithm exploits the recent advances of second order online learning techniques~\citep{dredze2008confidence}, trying to select the most confident weights while keeping the distribution close to the non-truncated distribution. It achieves highly competitive learning accuracy even compared with state-of-the-art batch FS methods.

The rest of this paper is organized as follows: Section 2 reviews related work; Section 3 presents the proposed method in detail; Section 4 discusses our empirical studies; and finally Section 5 draws our conclusions. More extensive results are also included in the appendix section due to space limitation.

\section{Related Work}
\label{sec:1}
Our work is related to feature selection and online learning. We review related work in each below.

Feature selection methods have been extensively studied in literature~\citep{kohavi1997wrappers,liu2005toward,zhao2013similarity},
which can be roughly grouped into three categories: \emph{Filter}, \emph{Wrapper}, and \emph{Embedded} methods. \emph{Filter} methods rely on
characteristics of data such as correlation, distance, and information gain without assuming specific classifiers~\citep{yu2003feature}.
Unlike the \emph{Filter} methods that ignore the effect of selected features on the performance of the induction algorithm, \emph{wrapper} methods employ a
predetermined classifier to evaluate the quality of selected features~\citep{kohavi1997wrappers}. It searches for a subset of
features and then evaluates their classification performances repeatedly. They often yield better performance for the chosen classifier, but are computationally intensive. \emph{Embedded} methods integrate feature selection into the model training process~\citep{xu2009non}, aiming to trade off between efficiency of \emph{filter} methods and predictive accuracy of \emph{wrapper} methods. However, their selected features might not be suitable for other classifiers.

Many studies have attempted to address online FS in diverse ways. Some aim to handle streaming features arriving sequentially to the classifier \citep{glocer2005online,perkins2003online,wu2010online}. Although they follow the stream learning setting and return a trained model at each time step given the observed features, they assume all the training instances must be given as a prior, making it unrealistic for many online applications. Our work is more closely related to another online FS setting in~\citep{wang2014online} that follows online learning methodology
by assuming training data arrives sequentially. Despite its considerable advantages in efficiency and scalability over batch FS methods,
it remains slow when being applied to large-scale FS tasks with ultra-high dimensionality.

Our work is also related to online learning in machine learning literature~\citep{crammer2006online,hoi2014libol}, where a variety of online algorithms have been proposed, ranging from classical first-order algorithms (such as Passive-Aggressive learning~\citep{crammer2006online}) to recent second-order algorithms~\citep{crammer2009adaptive}. In general, these algorithms require to access and explore the full set of features. They are not directly applicable to online FS tasks for selecting a fixed number of active features. Another closely related online learning method is sparse online learning~\citep{duchi2011adaptive,langford2009sparse}, which aims to learn a sparse linear classifier from training data in high-dimensional space. Despite the extensive efforts, most of these works usually impose a soft constraint, such as $\ell_1$-regularization, onto the objective function for promoting sparsity, which do not directly solve an online FS task that requires a hard constraint on the number of active dimensions in the learned
classifier. In this paper, we explore recent advances of online learning techniques in both second-order online learning and sparse online learning for advancing the state of the art of online feature selection tasks.

\section{Online Feature Selection}

In this section, we present a novel online feature selection method. We first describe the problem setting and then briefly introduce the existing first-order online feature selection methods, followed by presenting the proposed second-order online feature selection method in detail.

\subsection{Problem Setting}

Without loss of generality, this paper first investigates the problem of online feature selection for binary classification tasks. Consider $\{(\x_t,y_t) | t = 1,\ldots,T\}$ be a sequence of training data instances received sequentially over the training process, where each $\x_t \in \R^d$ is a vector of $d$ dimensions and $y_t \in \{+1,-1\}$. Generally, an online learner will learn a classifier with the same dimensionality $\w \in \R^d$. In the setting of online feature selection, we need to select a relatively small number of elements in $\w$ and set the others to be zero. In other words, we impose the following constraint
\begin{displaymath}
	\|\w\|_0 \leq B,
\end{displaymath}
where $B$ is the predefined constant, and consequently at most $B$ features of $\x$ will be used for prediction. Specifically, at each time $t$, a learner receives an incoming
example $\x_t \in \mathbb{R}^d$, and then predicts its class label $\hat{y}_t \in \{-1,+1\}$ based on its current model, i.e., a linear weight vector $\w_t$, as $$\hat{y}_t = \text{sign}(\w_t\cdot \x_t).$$ After making the prediction, the true label $y_t \in \{-1,+1\}$ will be revealed, and the learner then can measure the loss $l_t(\w_t)$ suffered with respect to $(\x_t,y_t)$, which is the difference between the prediction outcome and the true label. At the end of each iteration, the learner will update the weight vector $\w_t$ according to some learning rules. Throughout the paper, we assume $\|\x_t\| \leq 1, t = 1,\ldots, T$.

\subsection{First-order Online Feature Selection}

One of most straightforward approaches to online feature selection is to apply the Perceptron algorithm via truncation (PET) \citep{wang2014online}. Specifically, at each step, the classifier first predicts the label $\hat{y}_t$ with $\w_t$. If $\hat{y}_t$ is correct, then $\w_{t+1}=\w_t$; otherwise, the classifier will update $\w_t$ by Perceptron rule to obtain $\hat{\w}_{t+1}=\w_t+\eta_t y_t\x_t$, which will be further truncated by keeping the
largest $B$ absolute values of $\hat{\w}_{t+1}$ and setting the rest to zero. The truncated classifier, denoted by $\w_t^B$ or $\w_{t+1}$, will be used to predict the next observation.

As analyzed in~\citep{wang2014online}, the above simple approach does not work well in practice. In particular, it cannot guarantee a small number of mistakes since it fails to ensure the numerical
values of truncated elements are sufficiently small, thus leading to a nontrivial loss of accuracy. Consequently, the authors in \citep{wang2014online} proposed a novel first-order online feature selection
scheme (FOFS) by exploring online gradient descent with a sparse projection scheme before truncation, which guarantees the resulting classifier $\w_t$ to be restricted into an $\ell_1$-ball at each step. 
Algorithm~\ref{alg:pe-sp} shows the details of their first-order OFS algorithm.
\begin{algorithm}
	\caption{FOFS: First-order OFS via Sparse Projection}\label{alg:pe-sp}
	\begin{algorithmic} [1]
		\STATE\textbf{Input}: $B$,$\eta$
		\STATE Following the similar framework as PET but use constant learning
		rate $\eta$
		\STATE $\Tilde{\w}_{t+1} = (1-\lambda\eta)\w_t + \eta y_t\x_t$
		\STATE $\hat{\w}_{t+1} =
		\min\{1,\frac{\frac{1}{\sqrt{\lambda}}}{\|\Tilde{\w}_{t+1}\|_2}\}\Tilde{\w}_{t+1}$,  where $\lambda$ is a regularization parameter
		\STATE $\w_{t+1} = \mathrm{Truncate}(\hat{\w}_{t+1},B)$
	\end{algorithmic}
\end{algorithm}

\subsection{Second-order Online Feature Selection}

A key limitation of the above online feature selection algorithms is that they only exploit the first-order information of the weight vector during the online feature selection process, which may lead to the loss of potentially informative features. To overcome the limitation, we propose a second-order online feature selection method by exploring the recent advances of second-order online learning techniques.

The Confidence-Weighted (CW) method~\citep{dredze2008confidence} assumes the weight vector of the linear
classifier follows a Gaussian distribution $\w\sim \Nn(\Mu,\Sigma)$. Confidence of weights are represented by diagonal elements in covariance matrix $\Sigma_{j}$. The smaller $\Sigma_{j}$, the more
confidence we have in the mean value of weight $\mu_j$. Before observing any samples, all the weights are of the same confidence or uncertainty. In the CW learning process,
given an observed training example $(\x_t,y_t)$, CW makes an update by trying to stay close to the previous distribution and ensure that the probability of making correct prediction on $\x_t$ is larger than a threshold
$\eta$. The solution for the update can be cast into the following optimization:
\begin{equation}
	(\hat{\Mu}_{t+1},\Sigma_{t+1}) = \argmin_{\Mu,\Sigma}{D_{KL}(\Nn(\Mu,\Sigma), \Nn(\Mu_t,\Sigma_t))}  \quad s.t. \quad Pr_{\w\sim\Nn(\Mu,\Sigma)}[y_t(\w\cdot \x_t) \geq 0]  \geq \eta
	\label{equ:prob}
\end{equation}

The proposed second order online feature selection algorithm SOFS takes another
step with similar idea to CW. With the goal to reduce the damage to
classification ability while selecting features, SOFS tries to stay close to
the updated distribution and ensure the $L_0$ norm is less than $B$.
The updated weights $\hat{\w}_{t+1}$ in equation~\eqref{equ:prob} follows
the distribution $\hat{\w}_{t+1} \sim \Nn(\hat{\Mu}_{t+1},\Sigma_{t+1})$. SOFS is cast into the following optimization:
\begin{eqnarray}
	\Mu_{t+1} = \argmin_{\Mu}{D_{KL}(\Nn(\Mu,\Sigma_{t+1}),
	\Nn(\hat{\Mu}_{t+1},\Sigma_{t+1}))} \quad s.t. \quad \|\Mu\|_0 \leq B.
	\label{equ:sofs_formu_ori}
\end{eqnarray}

In SOFS, only diagonal elements of the covariance matrix $\Sigma$ are
considered. This is because maintaining a full covariance
matrix requires $O(d^2)$ memory space and $O(d^2)$ computational complexity,
which is impractical for handling large-scale ultra-high dimensional data.
By writing the KL divergence explicitly with the diagonal covariance matrix
assumption, the above optimization is equivalent to:
\begin{eqnarray}
	\label{equ:sofs_formu}
	\Mu_{t+1} = \argmin_{\Mu}{\frac{1}{2}(\Mu -
	\hat{\Mu}_{t+1})^T\Sigma_{t+1}^{-1}(\Mu -
	\hat{\Mu}_{t+1})} \quad s.t. \quad \|\Mu\|_0 \leq B.
\end{eqnarray}

Suppose the selected feature indexes of the optimal solution $\Mu^*$ to the
optimization are $s_1, s_2, \ldots, s_B$. Thus the rest feature weights with
indexes $s_{B+1}, \ldots, s_d$ are set to zero. The KL divergence is:
\begin{eqnarray}
	\label{equ:sofs_solu}
	KL(\Mu^*, \hat{\Mu}_{t+1}) =
	\Sigma_{i=1}^{B}{\Sigma^{-1}_{t+1,s_i}(\mu^*_{s_i} - \hat{\mu}_{t+1, s_i})^2}
	+ \Sigma_{i=B+1}^{d}{\Sigma^{-1}_{t+1,s_i}(\hat{\mu}_{t+1, s_i})^2},
\end{eqnarray}
where $\Sigma_{t+1,s_i}$ means the $s_i$-th diagonal element of the covariance
matrix at iteration $t$. As $KL(\Mu^*, \hat{\Mu}_{t+1})$ is the smallest among
all the possible $\Mu_{t+1}$, it can be drawn that:

\begin{compactitem}
	\item $\mu^*_{s_i} = \hat{\mu}_{t+1, s_i}, \forall i \in [1,B]$;
	\item  $\mu^*_{s_i} = 0, \forall i \in [B+1,d]$;
	\item $\Sigma_{t+1,s_i} \leq \Sigma_{t+1,s_j}, \forall i\in[1,B], j\in [B+1,d]$.
\end{compactitem}

Note that the covariance matrix represents the confidence of weights. The
above properties of the optimal solution indicate that the $B$ most
confidence features should be selected by exploiting the second-order
information of the classifier. Specifically, in the online learning process, when the loss for a training instance $(\x_t,y_t)$ is non-zero, we  update the weight vector only
for the most confident $B$ weight variables whose covariance values
$\Sigma_{j}$ are among the $B$ smallest, and all the other weights are set to
zero. By contrast, first order online feature selection algorithms select
important features based on the magnitudes of the classifier weights.

In this paper,  we adopt the Adaptive Regularization of Weights (AROW) algorithm~\citep{crammer2009adaptive} to solve the optimization problem in~\eqref{equ:prob}. It has been shown to be more robust in handling label noises than the original CW algorithms. The objective function of AROW is formulated as:
\begin{eqnarray}
	(\hat{\Mu}_{t+1},\Sigma_{t+1}) =
	\argmin_{\Mu,\Sigma}\big\{D_{KL}(\Nn(\Mu,\Sigma),
	\Nn(\Mu_t,\Sigma_t))
	+ \frac{1}{2\gamma}\ell_t(\Mu) + \frac{1}{2\gamma}\x_t^T\Sigma\x_t\big\},
	\label{equ:arow_obj}
\end{eqnarray}
where $\gamma>0$ is a regularization parameter. The problem in \eqref{equ:arow_obj} can be solved with closed-form solutions as follows:
\begin{eqnarray}
	\label{equ:arow_solu}
	\beta_t &= \frac{1}{\x_t^T\Sigma_{t}\x_t + \gamma} \quad \g_t^l  =
	-2\max(0, 1 - y_t \x_t^T\Mu_{t}) y_t\x_{t} \nonumber \\
	\Mu_{t+1} &= \Mu_{t} - \frac{1}{2}\beta_t\Sigma_{t}\g_t^l \quad \Sigma_{t+1}^{-1} = \Sigma_t^{-1} + \frac{\diag(\x_t^T\x_t)}{\gamma}
\end{eqnarray}

\vspace{-0.2in}
\subsection{Efficient SOFS Algorithms}

A common drawback with many existing second-order learning methods is the extra high computational cost incurred for exploiting the second-order information. In this section, we show that it is possible to 
devise a second-order OFS algorithm that is not only more effective but also considerably more efficient and scalable than the existing first-order approaches. 

Specifically, one of major time-consuming procedures in the above second-order feature selection method is to select top $B$ elements from an array of length $d$ (the diagonal vector of $\Sigma$ in
the second-order OFS). Instead of sorting all the weights at each step as in the previous study~\citep{wang2014online}, we propose a smart way to implement the 
proposed second-order online feature selection technique by employing a MaxHeap-based approach in exploiting the characteristics of SOFS, which can significantly reduce computational complexity to be linear with respect to the average number of nonzero features $m$ of each example, rather than the original full dimensionality $d$ ($d \gg m$). This makes it extremely fast and scalable when handling large-scale sparse high-dimensional data sets.

Before presenting the proposed algorithm, we first introduce the following proposition for the monotonic decreasing property of $\Sigma_t$, a property that is critical to the proposed algorithm.

\begin{prop}[monotonic decreasing] Given $\Sigma_t$ computed by \eqref{equ:arow_solu}, $\forall t$ and $\forall j\in[1,d]$, $\Sigma_{t+1,j} \leq \Sigma_{t,j}$.
\end{prop}
It is not difficult to verify the above by noticing ${\diag(\x_t^T\x_t)}/{\gamma}$ is
always non-negative. Using this important property, we can develop a fast algorithm for the second-order OFS method.

Specifically, we build a MaxHeap data structure to store the $B$ smallest diagonal values of covariance $\Sigma_t$.
The monotonic decreasing property of $\Sigma_t$ implies the heap limit should decrease monotonically. This leads to two major benefits in saving computational cost: (i) we do not need to check those unchanged elements to see
if they are smaller than the  heap limit; and (ii) when updating elements in the heap, only its child nodes need to be updated.

Algorithm~\ref{alg:fast_arow_fs} shows the details of the proposed fast algorithm for SOFS. Whenever a new feature arrives and its covariance changes, we proceed to update as follows:
\begin{compactitem}
\item If the corresponding covariance exists in the heap, adjust its position in the heap;
\item Check if it is smaller than the heap limit; if so, replace the root node of the heap by the current item and set
	the value of the original root node to be zero; otherwise,
\item Simply set the corresponding weight to zero.
\end{compactitem}

\begin{algorithm}
	\caption{SOFS: Fast Algorithm for Second-order OFS}\label{alg:fast_arow_fs}
	\begin{algorithmic}[1]
		\STATE\textbf{Input: $\gamma$, $B$}
		\STATE\textbf{Initialize:} $\Mu_1=0, \Sigma_1 = I$.  MaxHeap $H$ on $\Sigma_1$ with size $B$
		\STATE \textbf{for} {$t=1,\ldots, T$}
		\STATE \quad \textbf{if} $l_t(\Mu) = \max(0, 1 - y_t (\Mu\cdot \x_t))^2 > 0$
		\STATE \qquad Calculate $\beta_t$, $\g_t$ by \eqref{equ:arow_solu}.
		\STATE \qquad \textbf{for} {$j = 1, \ldots, d, \x_{t,j} \neq 0$}
		\STATE \quad \qquad ${\mu}_{t+1,j} = \mu_{t,j} - \frac{1}{2}\beta_t\Sigma_{t,j} g_{t,j}^l$, \quad $\Sigma_{t+1,j}^{-1} = \Sigma_{t,j}^{-1} + \frac{\x_{t,j}^2}{\gamma}$
		\STATE \quad \qquad \textbf{if} {$\Sigma_{t+1,j} \in H$}
		\STATE \qquad \qquad $\text{adjust } H \text{ to maintain the MaxHeap}$
		\STATE \quad \qquad \textbf{elseif}  {$\Sigma_{t+1,j} < H_{min}$}
		\STATE \qquad \qquad replace $H_{min}$ by $\Sigma_{t+1,j}$ and set the weight value of the original root node to be zero
		\STATE \qquad \qquad $\text{adjust } H \text{ to maintain the MaxHeap}$
		\STATE \quad \qquad \textbf{else}
		\STATE \qquad \qquad ${\mu}_{t+1,j} = 0$
		\STATE\textbf{Output: weight vector $\Mu_T$ and confidence $\Sigma_T$}
	\end{algorithmic}
\end{algorithm}

\subsection{Analysis of Time and Space Complexity}\label{sec:complexity}

The above proposed technique significantly improves the efficiency of existing online feature selection techniques. We now analyze the computational complexity of the above algorithms.

Let us denote by $d$ the dimensionality of the weight vector, and $m$ the average number of nonzero features of each sample.
For PET, each updating step has to calculate the loss ($O(m)$), update the
model ($O(m)$), calculate absolute value of the model ($O(m)$), find the largest $B$ elements according
to their absolute values and then set the rest $d-B$ to zero ($O(d + d\log B)$). The overall computational complexity of PET at every step is $O(3m + d + d\log B)$. FOFS
is similar to PET, with an extra normalization and sparse projection. The
extra complexity is $O(2d)$. Computational cost of calculating absolute value
is also increased to $O(d)$. Thus, the complexity of FOFS is $O(2m + 4d + d\log
B)$, which is much more computationally expensive for high dimensional data.
Our SOFS only needs to calculate the loss ($O(m)$), update weight vector and the covariance ($O(2m)$), and adjust the heap ($O(m\log B)$).
The computational complexity of SOFS at each step is reduced to $O(m\log B+3m)$, making it far
more efficient and scalable when handling ultra-high dimensional sparse data where $m\ll d$ and $B\ll d$. Even in the worst case where $m \approx d$, our SOFS with complexity $O(d\log B+3d)$ is still more efficient than PET ($O(4d + d\log B)$) and FOFS ($O(6d + d\log B)$), where the improvement even only a constant can still save lots of training time for ultra-high dimensional data.

For space complexity, we only consider the space required by the classifiers. Storages for data loading implementation are excluded here. Both PET and FOFS require to keep the weight vector $\w$ and its absolute vector $\v$ in memory, and thus have space complexity $O(2d)$. SOFS also has space complexity $O(2d)$ for keeping the weight vector and the diagonal elements of confidence matrix $\Sigma$ in memory. Thus, SOFS shares the same space complexity as the first-order online FS algorithms.

\section{Experiments}

In this section, we conduct extensive experiments to evaluate how the number of
selected features affects the test accuracy and the training efficiency of
different feature selection algorithms on both synthetic and real data on a large scale. We also evaluated the proposed algorithm on several public available medium-scale datasets. The results are shown in the supplementary material.

\subsection{Experimental Setup}

For the family of online feature selection algorithms, we only run each
algorithm by a single pass through the training data if without explicit
indication. We compare the proposed algorithm with a set of state-of-the-art algorithms including both online and batch feature selection as follows:
\begin{compactitem}
\item PET: the baseline of OFS by Perceptron with truncation \citep{wang2014online};
\item FOFS: the state-of-the-art first-order OFS via sparse projection \citep{wang2014online};
\item mRMR: minimum Redundancy Maximum Relevance Feature Selection, a state-of-the-art
	batch feature selection method~\citep{peng2005feature}.
\item Liblinear: a famous library for large linear classification~\citep{fan2008liblinear}. We adopt $l$1-SVM for the \emph{Embedded} feature selection in our experiments.
\item FGM: a batch \emph{Embedded} feature generating method~\citep{DBLP:journals/jmlr/TanTW14}.
\end{compactitem}

For online algorithms, we use hinge loss as the loss function. A five-fold cross validation is conducted to identify the optimal parameters. The experiments were conducted over 10 times with a random permutation of a dataset. For $l$1-SVM in liblinear, we
tune parameter $C$ to select different number of features. For FGM, we follow the settings
in~\citep{DBLP:journals/jmlr/TanTW14} and set $C = 10$ for simplicity. For mRMR, we
first select a specific number of features and then use the
Perceptron to train a classifier. We
exploited the advantage of online learning that processes data sequentially and
implemented the program with two parallel threads, one for data loading
and the other for learning. All experiments were conducted on a PC with Intel i7 CPU @ 3.3 GHz, 16 GB RAM \footnote{The source codes for our experiments will be released after the paper is published.}.

\subsection{Experiments on Synthetic Data}
The goal of this set of experiments is to generate synthetic data with ultra high dimensionality in order to examine different aspects of our algorithm in an effective way.

\begin{table}[htb]
\vspace{-0.4in}
	\caption{Summary of synthetic data (``K",``M",``B" are thousand, million, and billion, respectively.)}
	\label{tbl:synthetic_data}
	\center
	\begin{threeparttable}
		\begin{tabular}{ccccccc}
			\hline\noalign{\smallskip}
			DataSet      &  \#Train & \#Test &Dim & IDim\tnote{1} & NDim\tnote{2} &\#Feat\\
			\noalign{\smallskip}\hline\noalign{\smallskip}
			$\X_1$ & 100K        & 10K     & 20K  & 200 & 400     & 60M\\
			$\X_2$ & 1M     & 100K    & 1B     & 500 & 500  & 1B\\
			\noalign{\smallskip}\hline
		\end{tabular}
		\begin{tablenotes}
		\item[1] IDim is the dimension of informative features per instance
		\item[2] NDim is the dimension of noise features per instance
		\end{tablenotes}
	\end{threeparttable}
\vspace{-0.1in}
\end{table}

\textbf{Synthetic Data.} We follow the settings of FGM and generate two types of synthetic data, namely $\X_1
\in \R^{100K\times 20K}$ and $\X_2 \in \R^{1M\times 1B}$ to test efficacy, efficiency, and scalability of the algorithms for binary classification. Each entry is sampled from the $i.i.d.$ Gaussian distribution $\Nn(0,1)$. To simulate real data, each sample is a sparse vector. The numbers of informative features for the two datasets are $200$ and $500$ respectively. For each sample, we
randomly select $400$ dimensions for $\X_1$ and $500$ dimensions for $\X_2$ as noise. To generate labels, we sample a weight vector $\w^*$ from the Uniform distribution $\U(0,1)$ as the groundtruth weights for features. The label of each sample is determined by $y = sign(\w^*\cdot \x^*)$, where $\x^*$ is a sample without noise. 
Table~\ref{tbl:synthetic_data} summarizes the synthetic datasets.

\begin{figure}[htb]
\vspace{-0.1in}
	\subfigure[Test Accuracy]{ \includeMyGraphicX{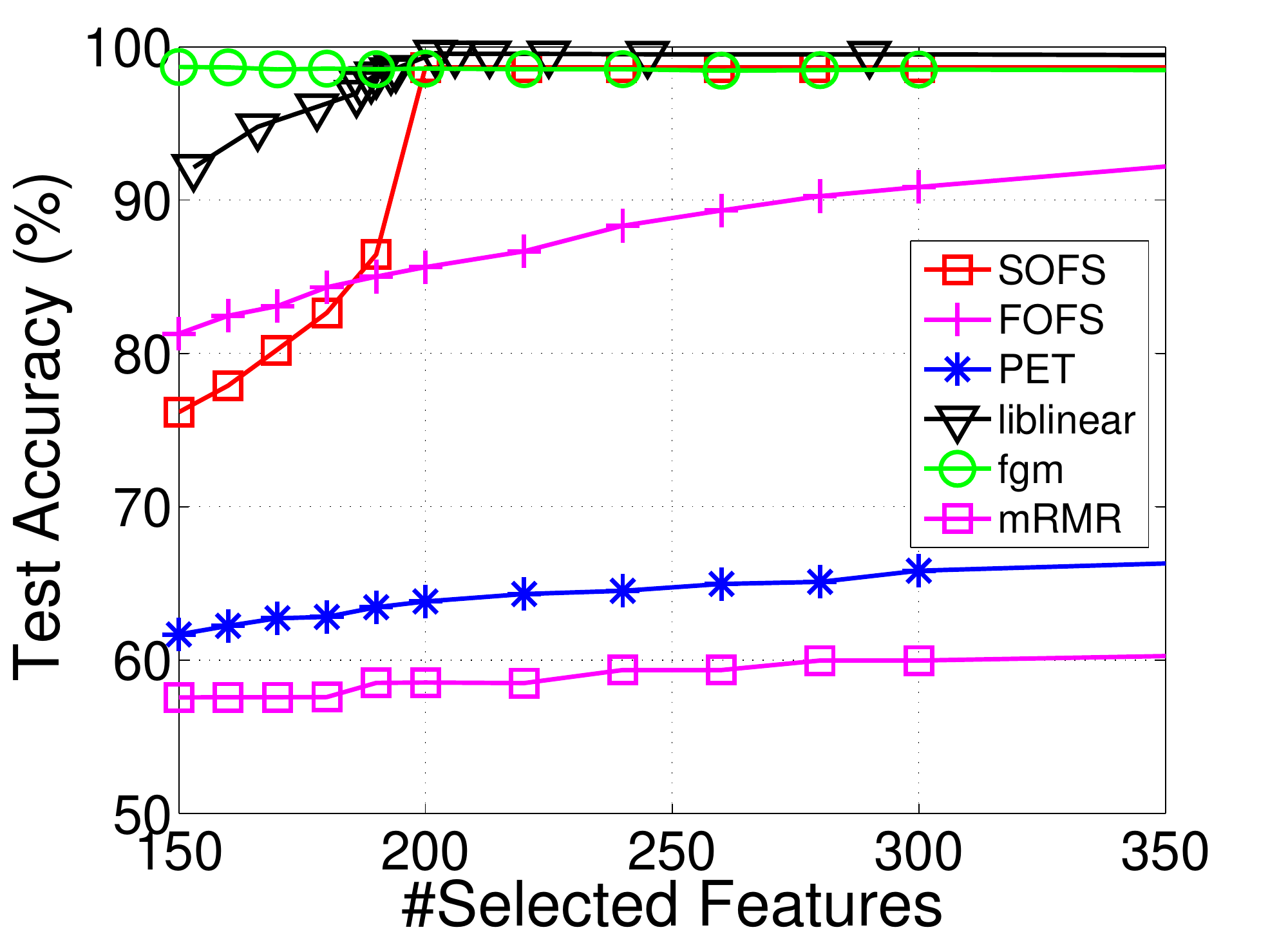} }
	\subfigure[Time Cost]{ \includeMyGraphicX{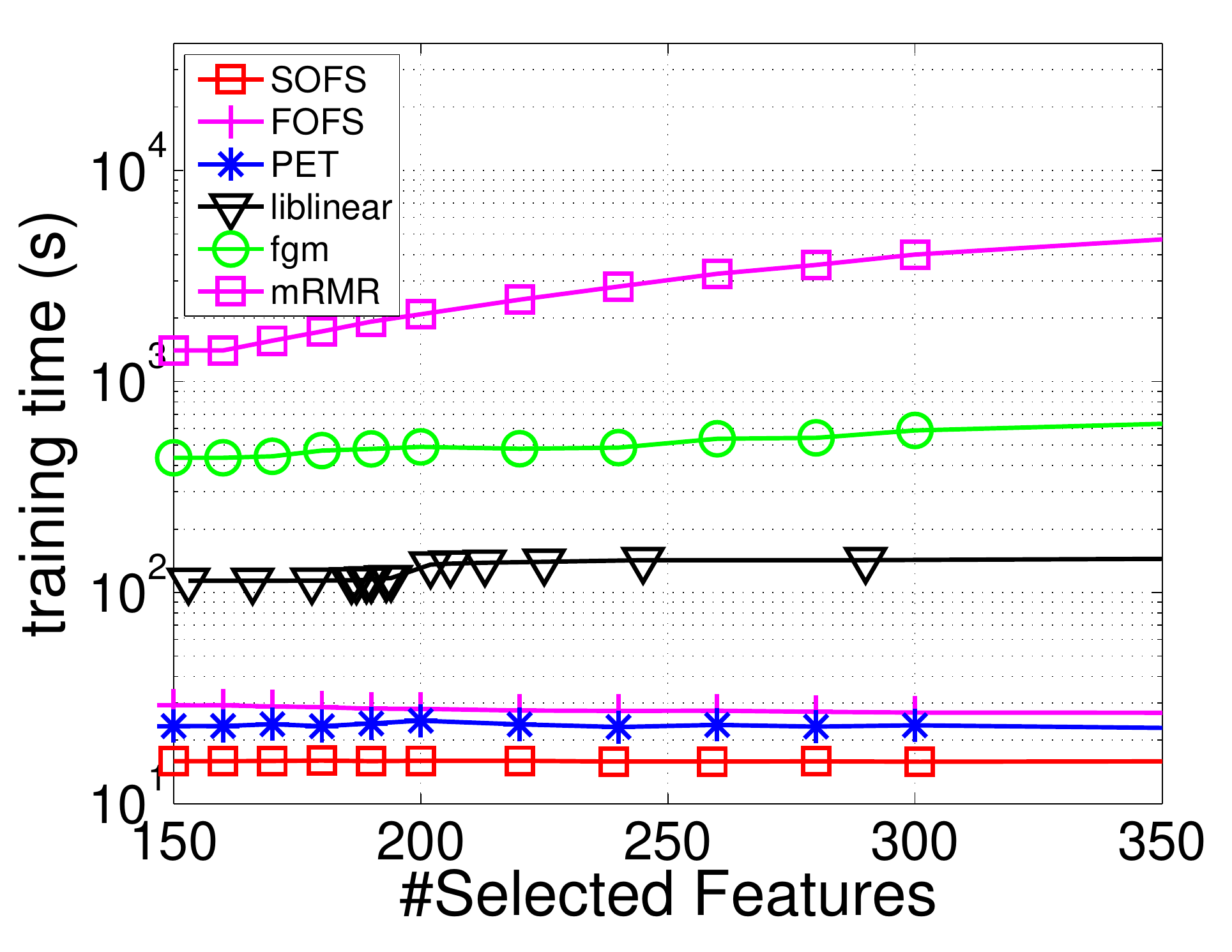}}\vspace{-0.1in}
	\caption{Evaluation of SOFS on the synthetic dataset $\X_1$}
	\label{fig:synthetic}
\vspace{-0.1in}
\end{figure}

Figure~\ref{fig:synthetic} shows the comparisons of accuracy and time cost.

\textbf{Accuracy.} According to Figure~\ref{fig:synthetic}(a), the proposed algorithm outperforms other online feature selection algorithms, showing its
efficacy in exploiting informative features. SOFS is superior to FOFS and PET significantly when the number of selected features exceeds
the number of informative features. mRMR performs the worst, similar to the observations in~\citep{wang2014online}. Batch learning algorithms are
superior to online algorithms when number of features is very limited. However, SOFS reaches the best and is comparable to batch feature selection algorithms when the number of selected features exceeds the number of informative feature (200 in $\X_1$). To conclude, the proposed algorithm is able to identify the groundtruth geometry of the data. 

\textbf{Time Cost.} Although batch FS algorithms are often more effective, they are significantly slower than online FS algorithms. Among the algorithms, our SOFS can achieve comparable test accuracy as batch FS algorithms with the lowest time cost (only a few seconds). By contrast, liblinear is 10 times slower and FGM is more than 1,000 times slower than SOFS on the dataset. Among online FS algorithms, our method has the best accuracy but requires the least time cost.


\textbf{Scalability on Ultra-High Dimensional Data.} Due to the ultra-high dimensionality and billion-scale features of $\X_2$, we found that it would have to take days to run the existing FS algorithms. We thus only compare SOFT with two variants using two kinds of online learning algorithms on full sets of features (by choosing $B=500$ for simplicity): Online Gradient Descent (OGD) and AROW \citep{crammer2009adaptive}. Note that these two baselines were also implemented efficiently using the same framework of SOFS with efficient data structure, but without doing feature selection. Table~\ref{tbl:sofs-ulta-high-D-syn} also shows the evaluation results on $\X_2$. 

\begin{table}[htpb]
\vspace{-0.2in}
	\caption{Evaluation on the ultra-high dimensional synthetic data $\X_2$}\label{tbl:sofs-ulta-high-D-syn}
	\center
	\small{
	\begin{tabular}{cccc}
		\hline\noalign{\smallskip}
		Algorithms & Time Cost (s) &  Accuracy & Sparsity (\%) \\
		\noalign{\smallskip}\hline\noalign{\smallskip}
		OGD & 266.76 & 99.30 &  83.52 \\
		AROW & 396.38 & 99.50  & 67.91 \\
		SOFS & 480.95 & \textbf{99.69} &  \textbf{99.995} \\
		\noalign{\smallskip}\hline
	\end{tabular}
	}
\vspace{-0.1in}
\end{table}

As seen from the results, SOFT has improved the test accuracy as compared to the two baselines without explicit FS, which verifies that removing irrelevant or noisy features can improve predictive performance. Not only with higher accuracy, SOFS also uses significantly less features (only $0.1\%$ as compared to $16\%$ by OGD and $32\%$ by AROW). In terms of time cost, SOFT took slightly more time cost due to the extra FS process, but only about 8 minutes to train a classifier on this dataset with billion-scale features. These encouraging results again validate that SOFT is efficient, scalable and effective in exploiting informative features on large-scale ultra-high dimensional data.


\subsection{Experiments on Large-scale Real-world Data Sets}

In this part, we evaluate the performance of the proposed SOFS algorithm for three large-scale text classification tasks, as shown in Table~\ref{tbl:large_dataset}. The first dataset `news` (for news group classification)
is high dimensional, the second `rcv1` (for text categorization) is relatively large scale, and the last one `url` (for suspicious url detection) is large scale and high dimensional.
In this experiment, for simplicity, we compare the proposed SOFS algorithm only with PET (due to its low time complexity) and FGM (due to its high accuracy).
\begin{table}[!htpb]
\vspace{-0.1in}
	\caption{Summary of large-scale real-world datasets in our experiments}
	\label{tbl:large_dataset}
	\center
	\small{
	\begin{tabular}{ccccc}
		\hline\noalign{\smallskip}
		DataSet & Feat Dim & Train No. &Test No. & Feat No. \\
		\noalign{\smallskip}\hline\noalign{\smallskip}
		news & 1,355,191 & 10,000  & 9,996  & 5,513,533 \\
		rcv1 & 47,152 & 781,265 & 23,149 & 59,155,144 \\
		url & 3,231,961 & 2,000,000    & 396,130 & 231,249,028\\
		\noalign{\smallskip}\hline
	\end{tabular}
	}
\end{table}

Table~\ref{tbl:comp_big} shows the experimental results of test accuracy and time cost of the three algorithms. We cannot show the results of FGM on ``url'' as it was too slow to run (took days to select 20\% features). We observe that the performance of SOFS is very close to that of FGM. Both PET and FGM are far more computationally expensive, with FGM  even more than an order of magnitude difference. The results further verify the significant advantage of SOFS on large-scale high-dimensional datasets.

\begin{table}[htpb]
	\caption{Evaluation on large-scale high-dimensional datasets ($\rho$ is the fraction of selected features).}\label{tbl:comp_big}
	\small{
	\begin{tabular}{c|lllll}
		\hline\noalign{\smallskip}
		Dataset & $\rho$ & 0.005 & 0.05 & 0.1 & 0.2\\
		\noalign{\smallskip}\hline\noalign{\smallskip}
		\multirow{3}{*}{news}& PET & 75.31$\%$(52.95s) & 71.71$\%$(46.76s) & 70.93$\%$(54.27s) & 76.55$\%$(61.76s) \\
		&SOFS &  78.48$\%$(\textbf{1.73s}) & 79.3$\%$(\textbf{2.12s})  & 79.36$\%$(\textbf{1.65s})  & 79.52$\%$(\textbf{1.45s})\\
		&FGM  & 79.3$\%$(75.40s)& 79.71$\%$(751.01s) & 79.72$\%$(2540.64s) & 79.63$\%$(7587.04s)\\
		\noalign{\smallskip}\hline\noalign{\smallskip}
		\multirow{3}{*}{rcv1} & PET & 80.14$\%$(140.15s)& 92.35$\%$(79.90s)& 93.71$\%$(82.43s) & 94.24$\%$(99.95s)\\
		&SOFS & 87.59$\%$(\textbf{14.71s}) & 93.65$\%$(\textbf{16.40s}) & 94.23$\%$(\textbf{15.62s}) & 94.61$\%$(\textbf{15.6s})\\
		&FGM & 94.58$\%$(501.4s)& 94.71$\%$(950.1s) & 94.76$\%$(1339.0s)& 94.81$\%$(2039.1s)\\
		\noalign{\smallskip}\hline\noalign{\smallskip}
		\multirow{2}{*}{url} & PET &  98.16$\%$(3816.8s)& 98.24$\%$(4750.4s) & 98.29$\%$(4957.3s) &	98.29$\%$(4878.7s)\\
		&SOFS & 98.40$\%$(\textbf{66.7s})  & 98.62$\%$(\textbf{68.6s}) & 98.66$\%$(\textbf{68.9s}) & 98.71$\%$(\textbf{67.1s}) \\
		\noalign{\smallskip}\hline
	\end{tabular}
	}
\end{table}

\section{Conclusions}

This paper addressed an open challenge of large-scale feature selection with large-scale ultra-high dimensional sparse data, and presented a novel scheme of Second-order Online Feature
Selection (SOFS). In contrast to the existing online FS algorithms whose computational complexity is linear with respect to the total feature dimensions, the
proposed new SOFS algorithm has a significantly lower computational complexity that is linearly dependent on the average number of nonzero features with each instance. We extensively evaluated empirical
performance of the proposed algorithm by comparing it with state-of-the-art online and batch feature selection algorithms on both synthetic and large-scale real
datasets. The promising results showed that our new method not only achieved highly competitive prediction accuracy, but also significantly improved computational efficiency, making our method practical for handling large-scale sparse data with ultra-high dimensionality.

\bibliographystyle{iclr2016_conference}

\section*{APPENDIX: More Experiments on Medium-Scale Real Data Sets}

In this appendix, we give more extensive experimental results of performance evaluations on a variety of medium-scale real-world datasets.

\subsection*{Overview of Medium-Scale Real Data Sets}

In this section, we evaluate the performance of online feature selection
algorithms on a number of medium-scale public benchmark datasets, as shown in Table~\ref{tbl:dataset}.
The datasets can be downloaded either from Feature Selection website of Arizona State
University\footnote{\url{http://featureselection.asu.edu/datasets.php}} or
SVMLin\footnote{\url{http://vikas.sindhwani.org/svmlin.html}} (for sparse datasets).

\begin{table}[!htpb]
	\caption{Medium-scale real datasets in experiments}
	\label{tbl:dataset}
	\center
	\begin{tabular}{ccccc}
		\hline\noalign{\smallskip}
		DataSet & Feat Dim & Train No. &Test No. & Feat No. \\
		\noalign{\smallskip}\hline\noalign{\smallskip}
		relathe & 4,322 & 1,000    & 427      & 87,352 \\
		pcmac & 7,510 & 1,000 & 946 & 55,470 \\
		basehock & 4,862 & 1,500    & 493 & 101,974\\
		ccat & 47,236 & 13,149 & 10,000 & 994,133 \\
		aut & 20,072 & 40,000 & 22,581 & 1,969,407 \\
		real-sim & 20,958& 50,000 & 22,309 & 2,560,340 \\
		\noalign{\smallskip}\hline
	\end{tabular}
\end{table}

\subsection*{Evaluation of Accuracy}

\begin{figure}[!t]
	\subfigure[relathe]{ \includeMyGraphicX{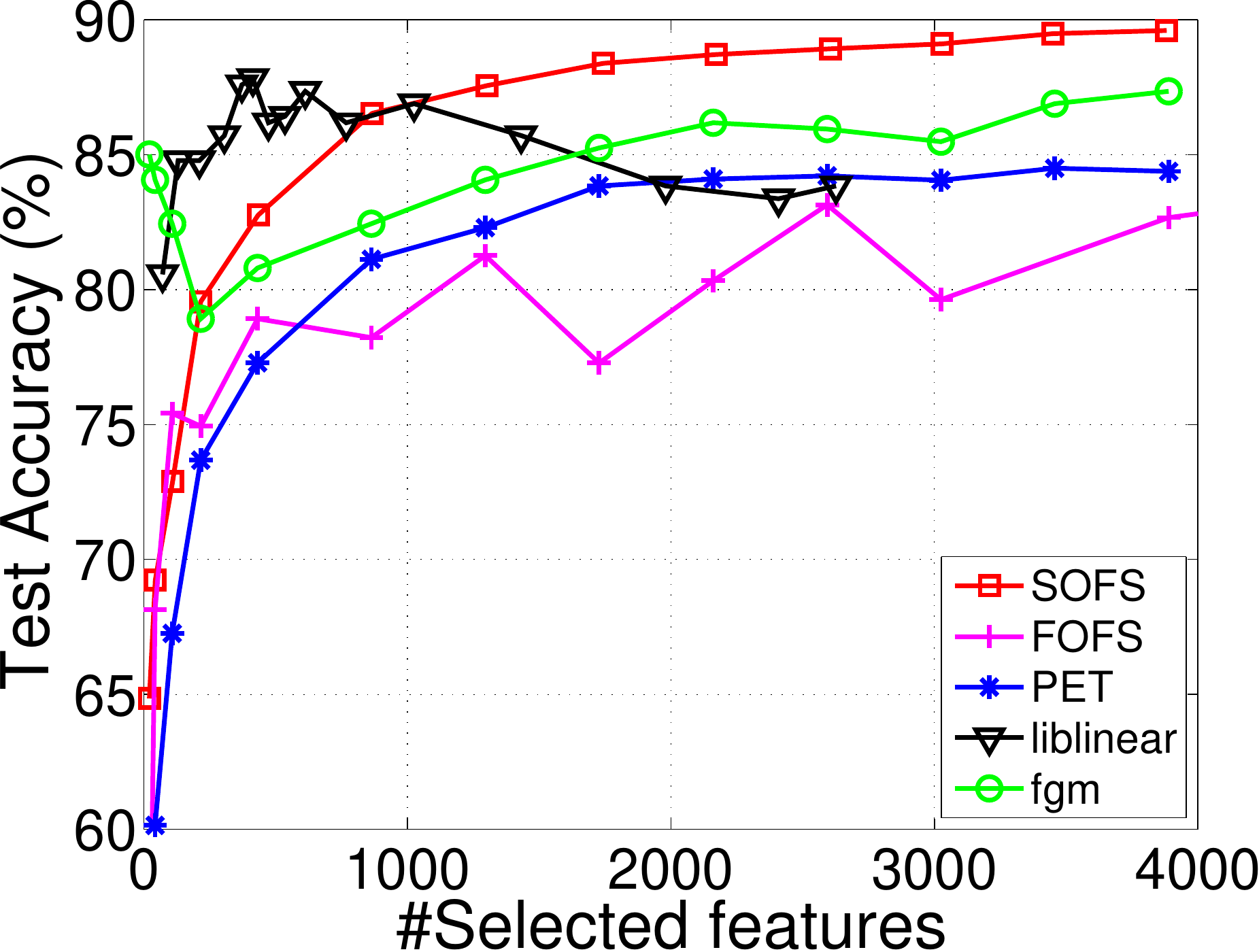}}
	\subfigure[pcmac]{ \includeMyGraphicX{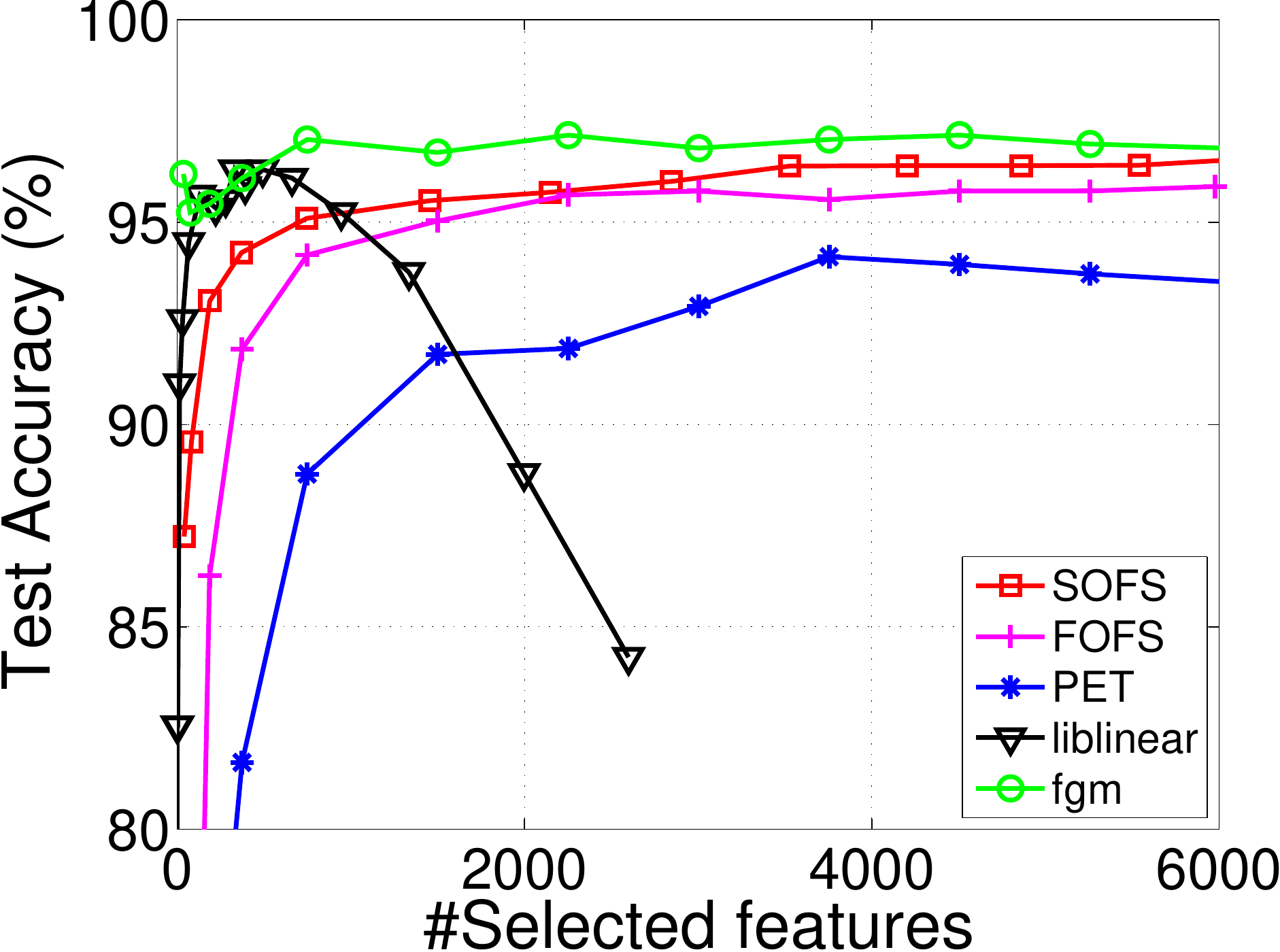}}\vspace{0.2in}
	\subfigure[basehock]{ \includeMyGraphicX{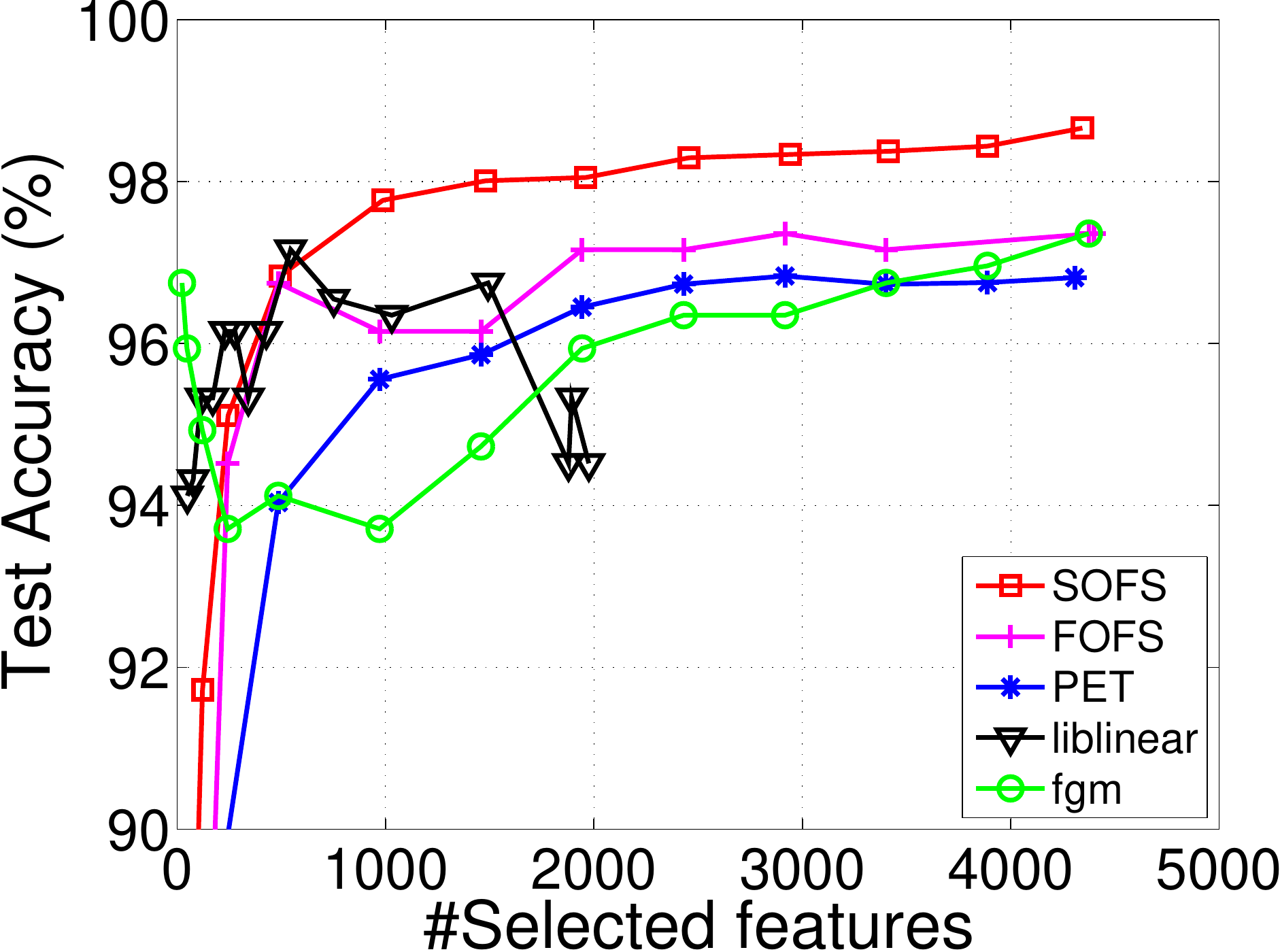}}
	\subfigure[real-sim]{ \includeMyGraphicX{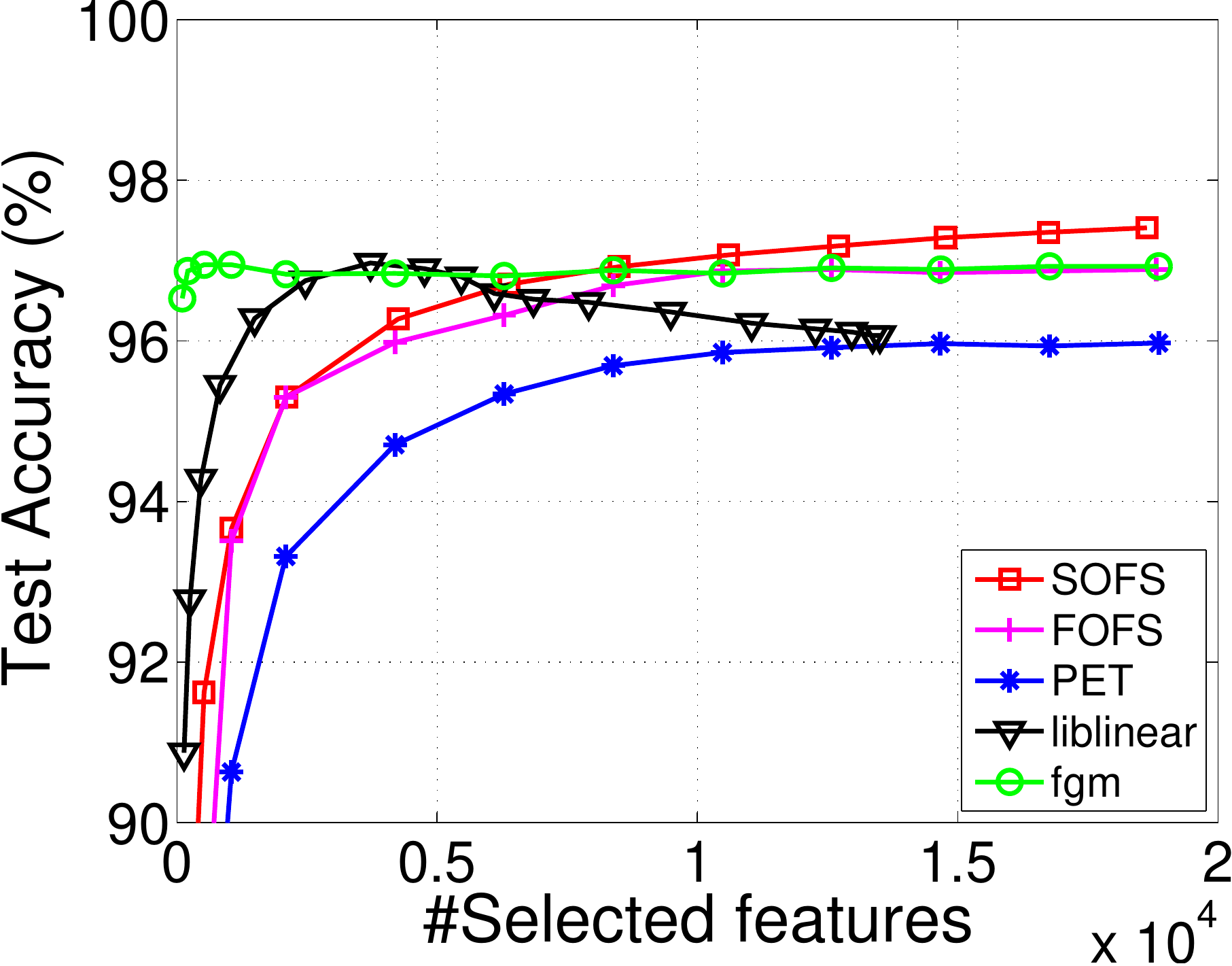}}\vspace{0.2in}
	\subfigure[ccat]{ \includeMyGraphicX{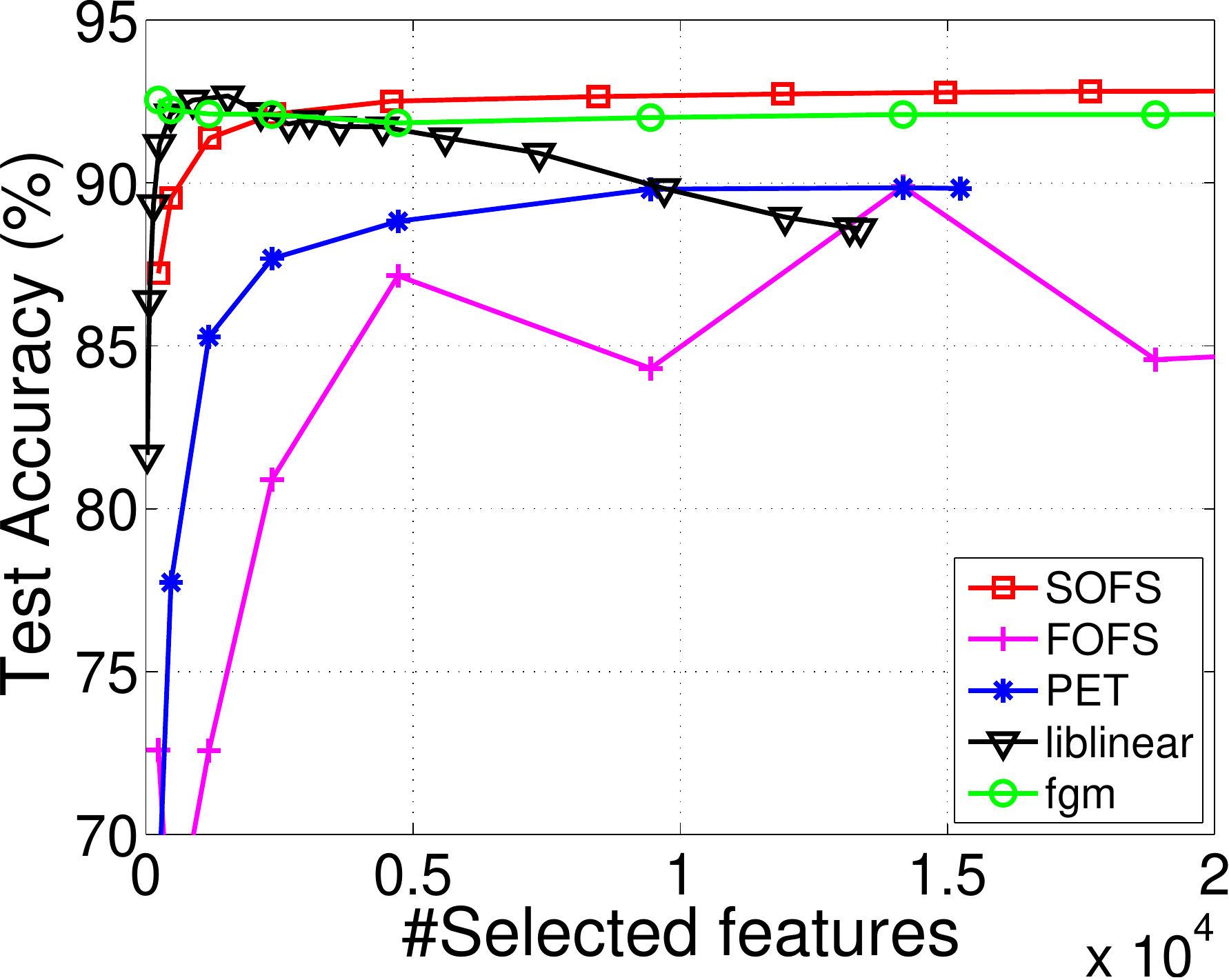}}\hspace{1.1cm}
	\subfigure[aut]{ \includeMyGraphicX{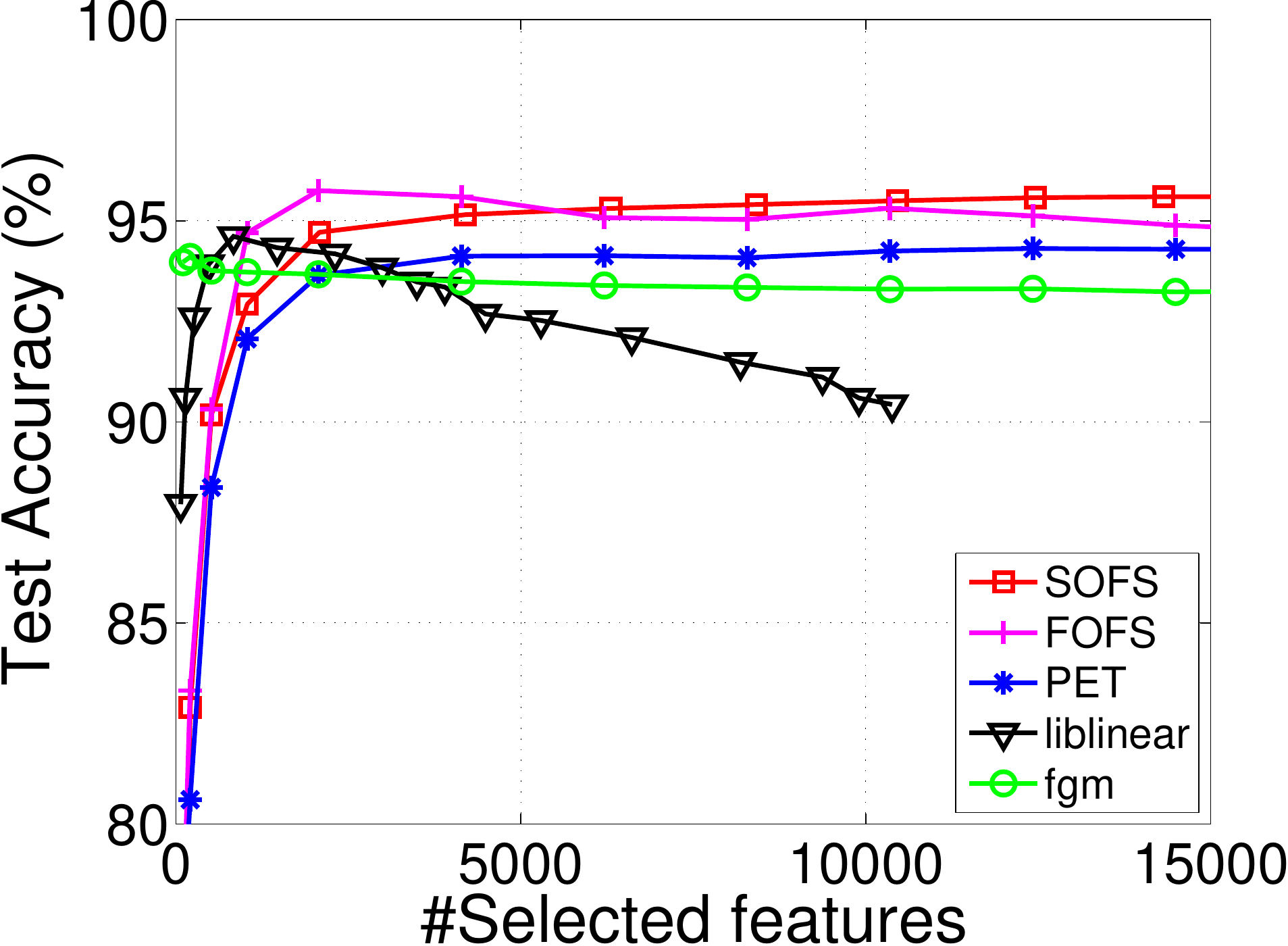}}
	\caption{Test Accuracy of Feature Selection Algorithms on Medium-scale real world data}
	\label{fig-online-medium-test-others}
\end{figure}

Figure~\ref{fig-online-medium-test-others} shows the test accuracy of different algorithms. By examining the online algorithms, we found that Perceptron (``PET") with a simple truncation does not work well, while
FOFS is much better than PET in most cases. However, we observe that performance of FOFS is not stable. The variance of FOFS is much larger than those of the other two online algorithms on half of the medium-scale datasets.
The proposed SOFS method is able to learn a more compact classification model. With the same number of selected features, SOFS is able to achieve the higher test accuracy results.

\begin{table}[!htpb]
	\caption{Comparison of SOFS with mRMR on medium-scale datasets}
	\label{tbl:comp_medium_batch}
	\center
	\begin{tabular}{ccccccc}
		\hline\noalign{\smallskip}
		Dataset & $B$ & 100 & 200 & 300 & 400 & 500 \\
		\noalign{\smallskip}\hline\noalign{\smallskip}
		\multirow{2}{*}{relathe} & mRMR & \textbf{74.19} & 77.87 &  78.92& 79.13 & 79.60 \\
		&SOFS & 71.38 & \textbf{78.81} &  \textbf{81.34} &  \textbf{82.39} & \textbf{82.91} \\
		\noalign{\smallskip}\hline\noalign{\smallskip}
		\multirow{2}{*}{pcmac} &mRMR & 87.95 &  90.34 &  89.93 &  91.49 &  91.10  \\
		&SOFS & \textbf{89.76} & \textbf{92.65} & \textbf{93.28} & \textbf{93.75} & \textbf{94.02}  \\
		\noalign{\smallskip}\hline\noalign{\smallskip}
		\multirow{2}{*}{basehock} &mRMR &  \textbf{93.78} & \textbf{95.15} &  95.03 &  95.25 &  94.89 \\
		&SOFS &  90.34 &  94.52 & \textbf{95.86} &  \textbf{96.41} & \textbf{96.68} \\
		\noalign{\smallskip}\hline\noalign{\smallskip}
		\multirow{2}{*}{ccat} &mRMR &  82.75 &  85.71 &  86.42 &  86.94 &  87.40 \\
		&SOFS &  \textbf{82.76} &  \textbf{86.35} &  \textbf{87.94} &  \textbf{89.00} &  \textbf{89.75} \\
		\noalign{\smallskip}\hline\noalign{\smallskip}
		\multirow{2}{*}{aut} &mRMR &  \textbf{92.41} & \textbf{93.87} & \textbf{94.09} &  94.59 &  94.55 \\
		&SOFS & 74.72 & 81.71 & 85.89 & \textbf{97.67} & \textbf{99.75} \\
		\noalign{\smallskip}\hline\noalign{\smallskip}
		\multirow{2}{*}{real-sim} &mRMR & \textbf{85.44} & \textbf{88.51} & \textbf{89.71} & \textbf{90.84} & \textbf{94.55} \\
		&SOFS & 83.29 & 86.77 & 89.38 & 90.71 & 91.59 \\
		\noalign{\smallskip}\hline\noalign{\smallskip}
	\end{tabular}
\end{table}

Besides, SOFS is comparable to batch FS algorithms when accuracy saturates with number of features. We find that FGM is able to perform well
with rather few features. Liblinear in this case shows a very interesting phenomenon in that the test accuracy first increases rapidly with more selected features, but after a certain stage
where the accuracy of other algorithms begins to saturate, the accuracy of Liblinear tends to drop considerably.
This implies that Liblinear may be more sensitive to irrelevant features or noises. 

We show the comparison of SOFS with mRMR separately in Table~\ref{tbl:comp_medium_batch} (as mRMR was only able to
output at most 500 selected features). From these results, we can observe that mRMR is better when the number of features is less. The accuracy of SOFS increases
quickly and surpasses mRMR with more selected features. This is consistent to the above results.  Note that mRMR is better than SOFS on the dataset
``real-sim''. In Figure~\ref{fig-online-medium-test-others}, all online FS algorithms fail to train a good model with only 500 features on ``real-sim''. Their performance
increases quickly and is expected to outperform mRMR with more features. The comparison again verifies the advantage of batch learning algorithms on
very small number of selected features. However, when more features are selected, the proposed online feature selection becomes more accurate than mRMR.

\subsection*{Evaluation of Time Cost}

\begin{figure}[!h]
	\subfigure[relathe]{ \includeMyGraphicX{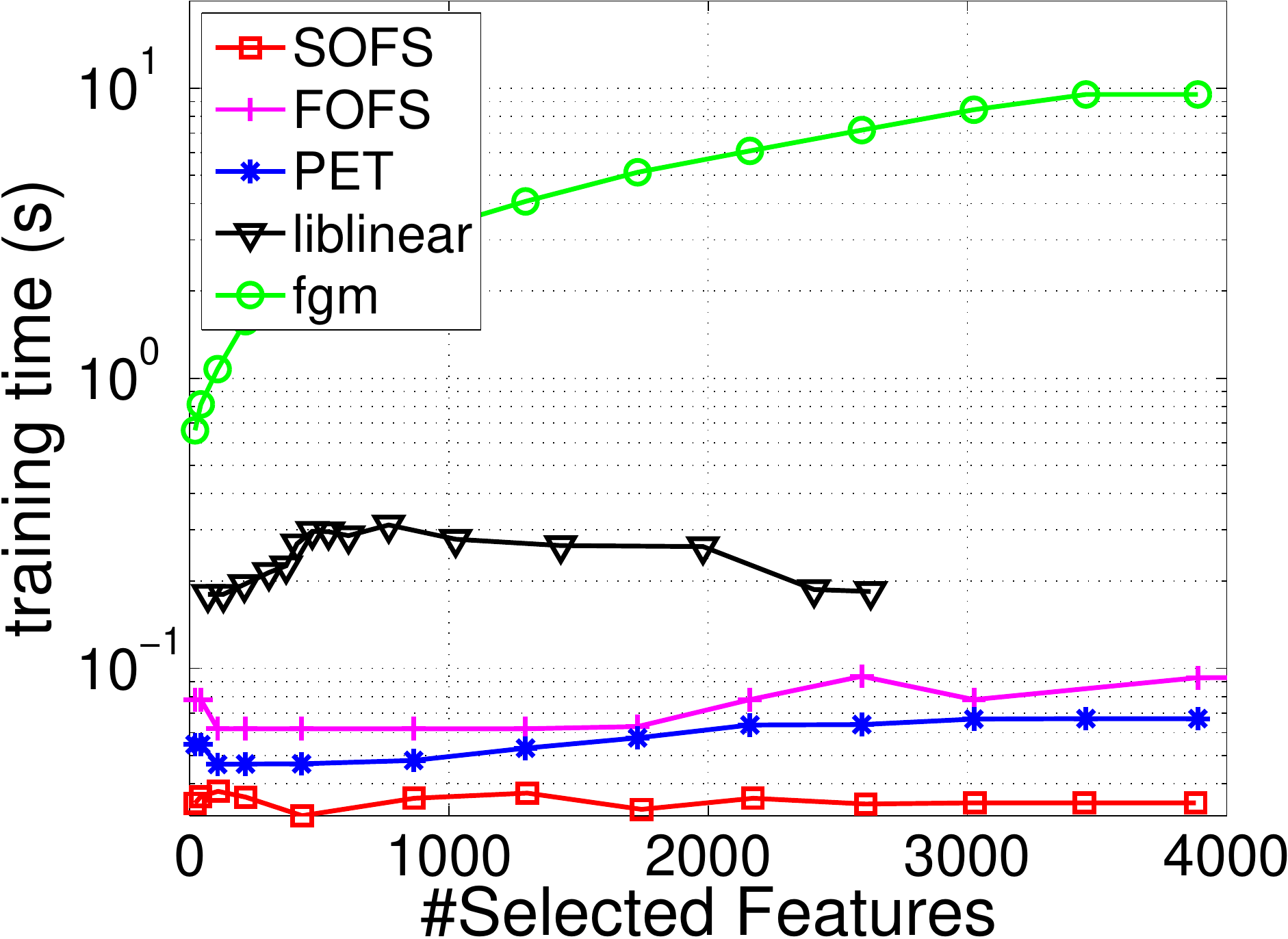}}
	\subfigure[pcmac]{ \includeMyGraphicX{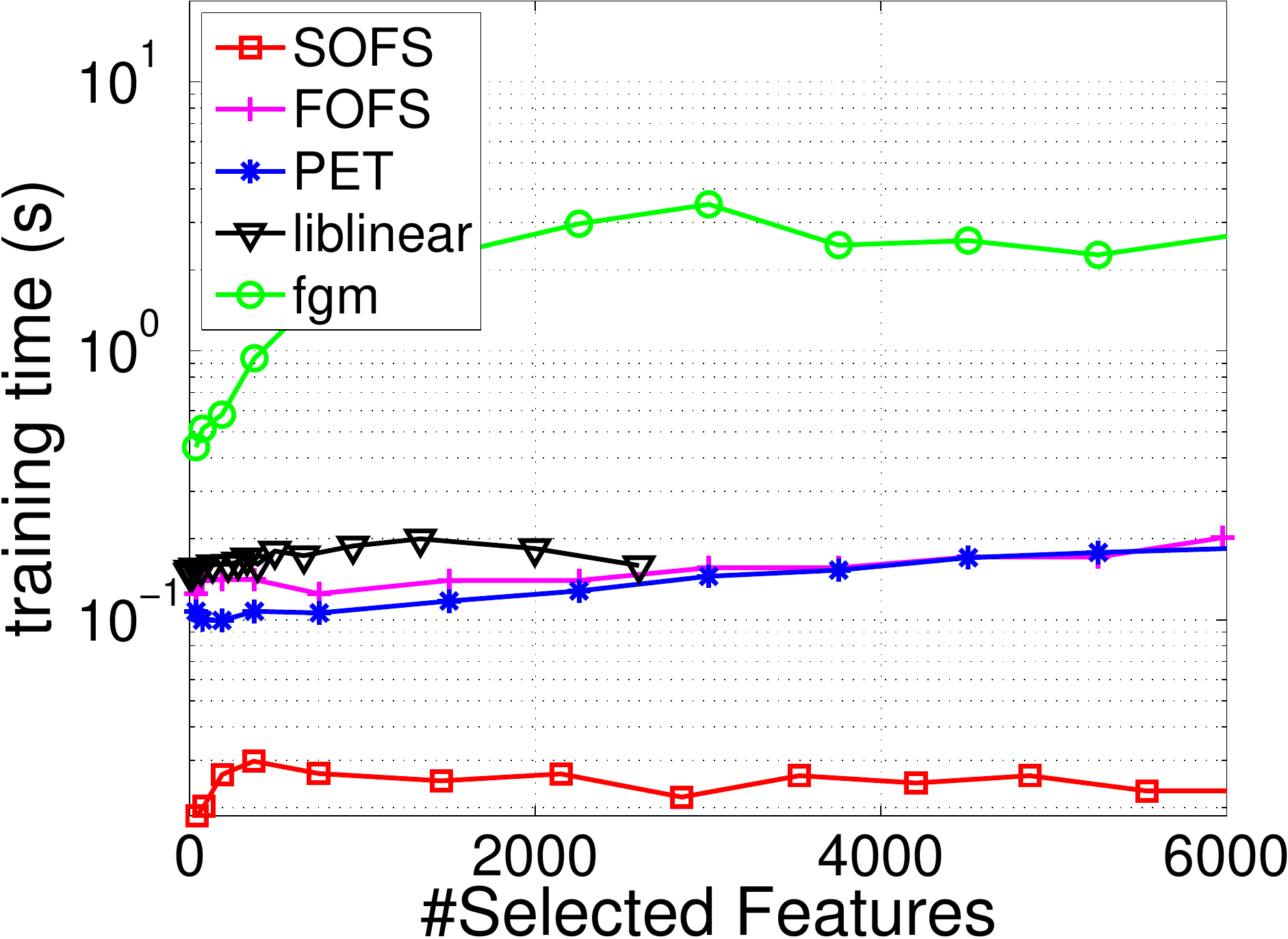}}\vspace{0.2in}
	\subfigure[basehock]{ \includeMyGraphicX{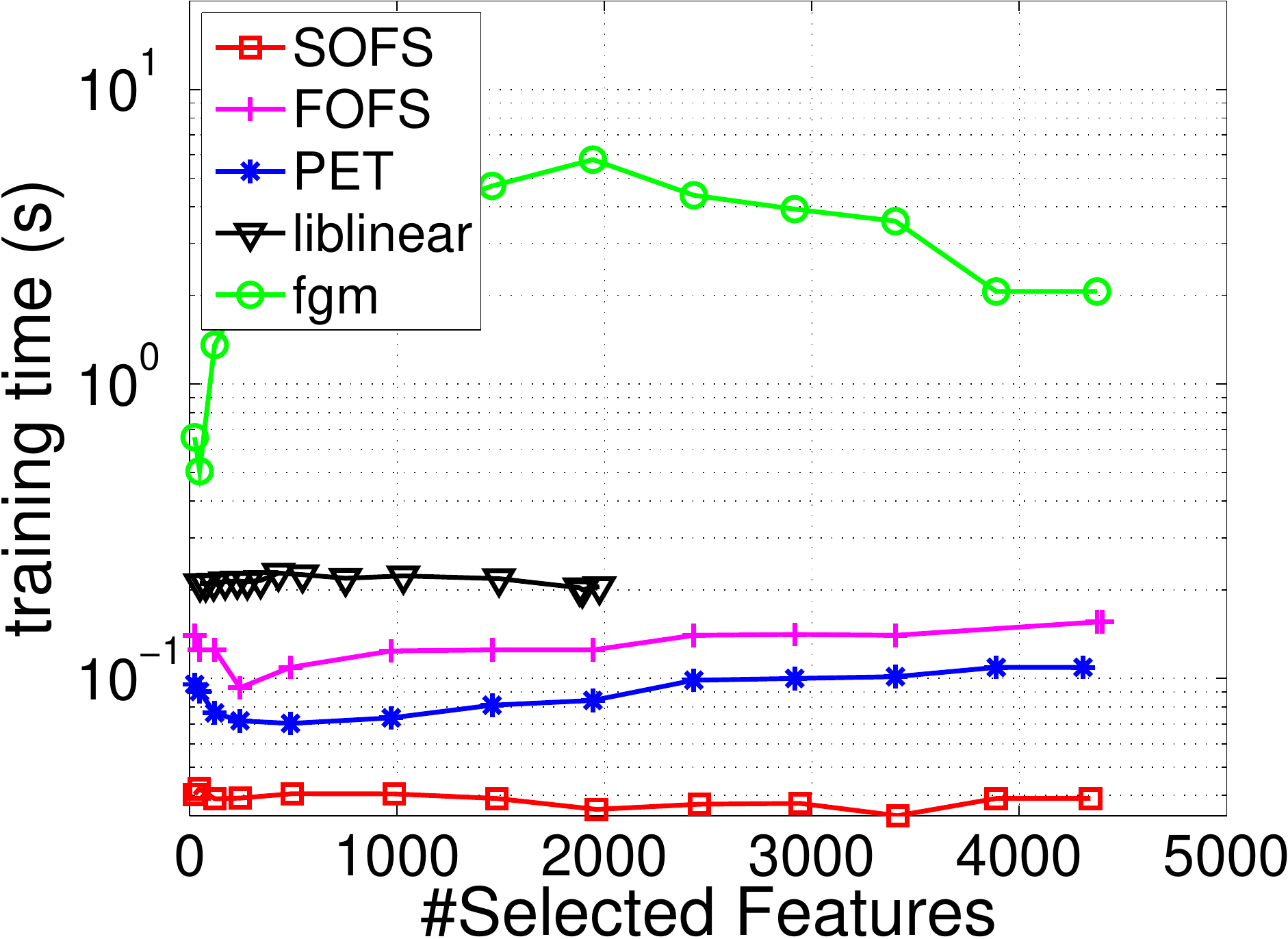}}
	\subfigure[real-sim]{ \includeMyGraphicX{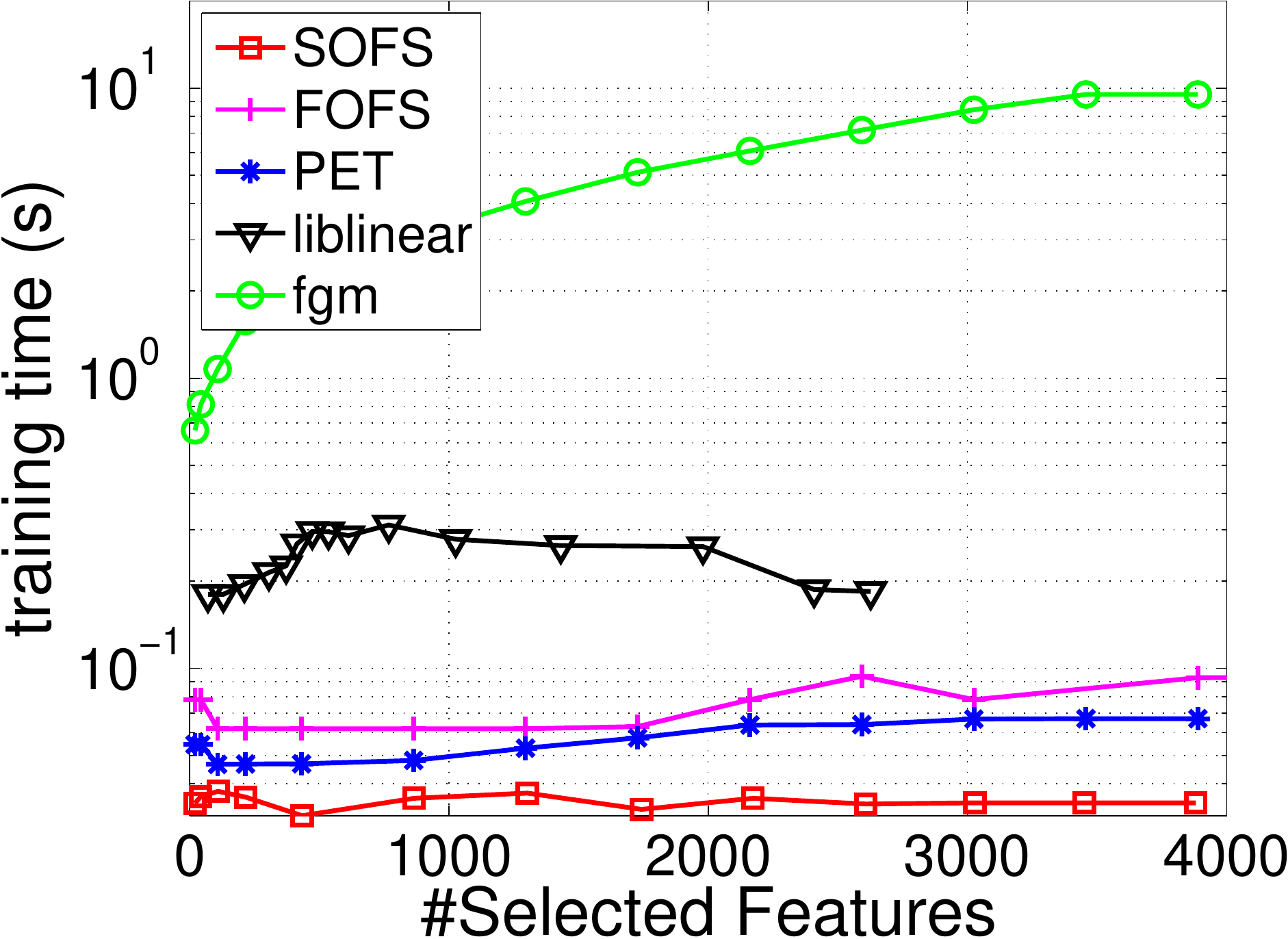}}\vspace{0.2in}
	\subfigure[ccat]{ \includeMyGraphicX{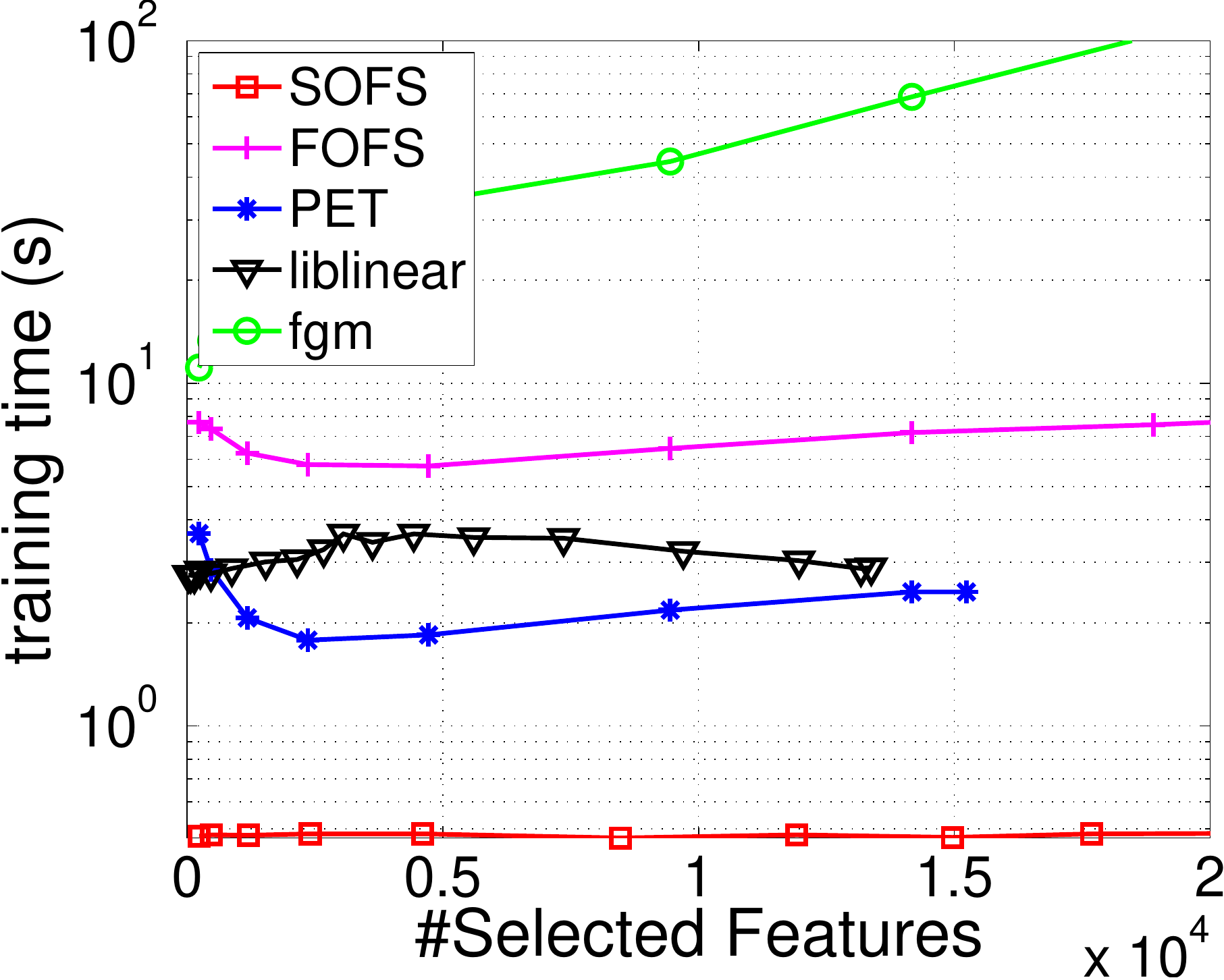}}\hspace{1.1cm}
	\subfigure[aut]{ \includeMyGraphicX{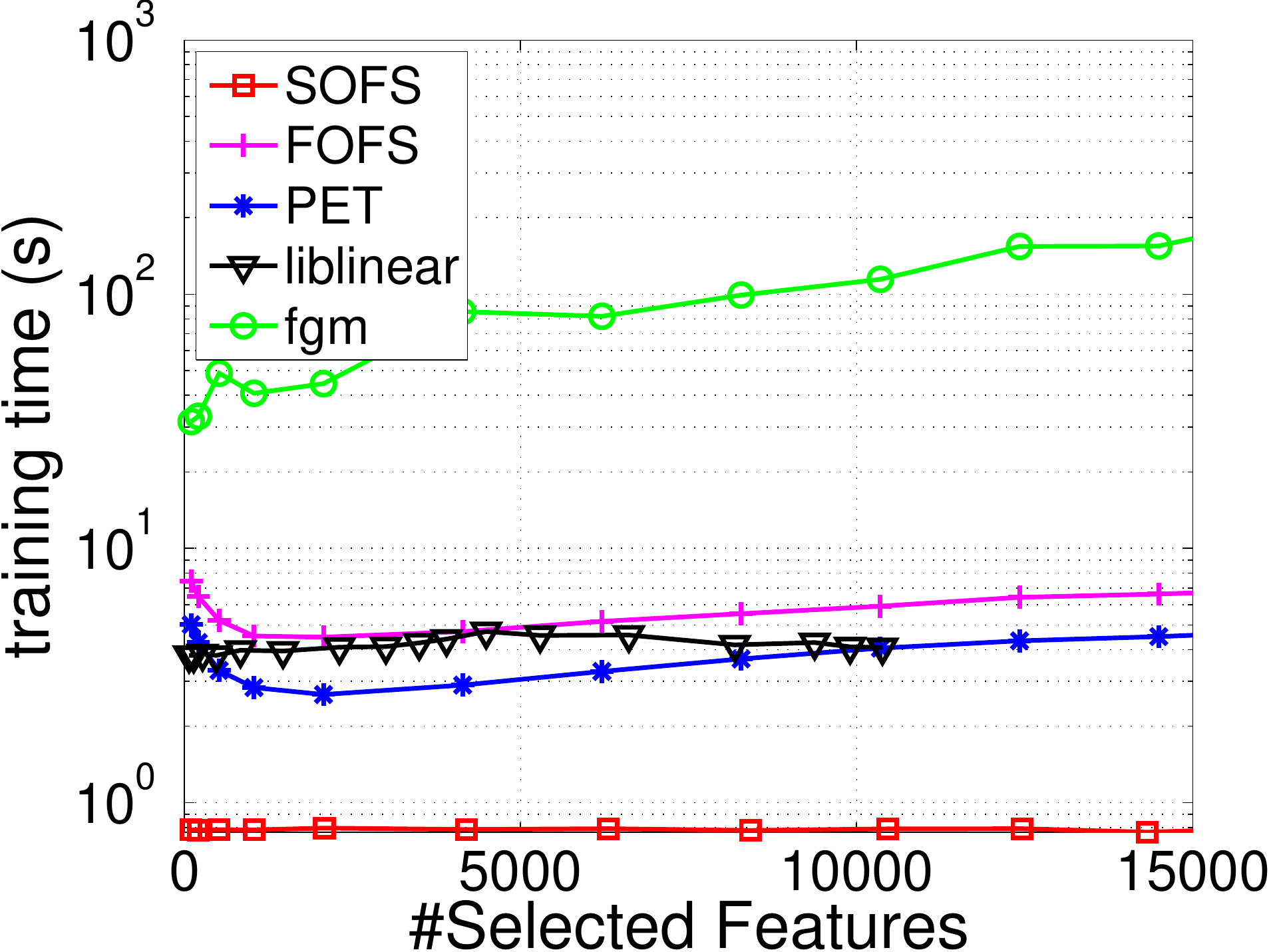}}
	\caption{Time Cost of Feature Selection Algorithms}
	\label{fig-online-medium-time-others}
\end{figure}

Figure~\ref{fig-online-medium-time-others} shows the time cost comparison of
feature selection methods on medium-scale data. First of all, we observe that FOFS took slightly higher time cost than PET despite achieving better accuracy. The extra time cost is more obvious when data dimensions get higher.  Further, we observe that the time cost first decreases and then increases with more selected features. This is due to the fact that when the number of selected features is too small, large number of mistakes are made and
the model has to update frequently. With more features, the prediction accuracy can be improved and thus less update is performed, resulting in the decreased time costs. Note the time costs on
the later three datasets, which are of relatively high dimension. It shows the great advantage of our proposed algorithm on high dimensional data. This is consistent with the
analysis in the paper that complexity of SOFS is linearly dependent on the number of non-zero features, while PET and FOFS are linearly dependent on the feature dimension.

\begin{table}[htb]
	\caption{Time Cost Comparison of SOFS with mRMR (\#features = 500)(seconds)}
	\label{tbl:time_mRMR}
	\center
	\begin{tabular}{ccccccc}
		\hline\noalign{\smallskip}
		Dataset & relathe &  pamac & basehock & ccat & aut & real-sim \\
		\noalign{\smallskip}\hline\noalign{\smallskip}
		SOFS& 0.03& 0.03& 0.04 & 0.48& 0.78& 1.24\\
		mRMR& 1733 & 1429 & 1584 & 2205 & 1486 & 1403 \\
		\noalign{\smallskip}\hline
	\end{tabular}
\end{table}

As to batch learning algorithms,  liblinear is
quite similar to first-order online algorithms, but is much more than that of
SOFS. Time cost of FGM is about an order of magnitude higher than liblinear. mRMR is the
most inefficient among all the algorithms. We show time cost of mRMR and SOFS
to select 500 features in TABLE~\ref{tbl:time_mRMR}. Even on the smaller dataset
``relathe'', it takes over 1,700 seconds to select 500 features, while SOFS
requres only 0.03 seconds. To conclude, SOFS is the most efficient one among all the algorithms in our experiments.

\end{document}